\documentclass[11pt,twoside,twocolumn,a4paper]{article}

\usepackage{cvww}
\usepackage{times}
\usepackage{epsfig}
\usepackage{graphicx}
\usepackage{amsmath}
\usepackage{amssymb}
\usepackage{verbatim}
\usepackage{lipsum}  
\usepackage{balance}
\usepackage{cancel}
\usepackage{xfrac}
\usepackage{color} 
\definecolor{mygreen}{rgb}{0.1,0.5,0}
\definecolor{myblue}{rgb}{0,0,0.6}

\newcounter{mytodo}

\usepackage[pagebackref=true,breaklinks=true,bookmarks=false]{hyperref}

\cvwwfinalcopy 

\def\cvwwPaperID{28} 

\ifcvwwfinal\fi

\begin{document}

\title{Robust Deformation Estimation in Wood-Composite Materials using Variational Optical Flow}

\author{Markus Hofinger, Thomas Pock\\
Institute of Computer Graphics and Vision (ICG)\\
Graz University of Technology\\
{\tt\small \{markus.hofinger, pock\}@icg.tugraz.at}
\and
Thomas Moosbrugger,\\
Rubner Holding AG\\
Kiens, Italy\\
{\tt\small{thomas.moosbrugger@rubner.com} }
}

\maketitle
\ifcvwwfinal\thispagestyle{fancy}\fi

\begin{abstract}
\label{sec:Abstract}
Wood-composite materials are widely used today as they homogenize humidity related directional deformations.
Quantification of these deformations as coefficients is important for construction and engineering and topic of current research  \cite{Moosbrugger2017_T1,Moosbrugger2018_T3}, but still a manual process.

This work introduces a novel computer vision approach that automatically extracts these properties directly from scans of the wooden specimens, taken at different humidity levels during the long lasting humidity conditioning process.
These scans are used to compute a humidity dependent deformation field for each pixel, from which the desired coefficients can easily be calculated.

The overall method includes automated registration of the wooden blocks, numerical optimization to compute a variational optical flow field which is further used to calculate  dense strain fields and finally the engineering coefficients and their variance throughout the wooden blocks.
The methods regularization is fully parameterizable which allows to model and suppress artifacts due to surface appearance changes of the specimens from mold, cracks, etc. that typically arise in the conditioning process.

\end{abstract}

\section{Introduction}
\label{sec:Intro}
Since this paper addresses a highly interdisciplinary research topic, a short introduction to humidity related wood deformations is given, before our and related work is presented.
\subsection {Deformation of wood w.r.t. humidity}
\label{sec:Intro_WoodProperties}
Wood, as orthotropic material, shows a humidity dependent deformation behavior that varies greatly in the different fiber directions.
Whilst the directions parallel to the fiber hardly change at all, the orthogonal radial directions show strong changes \cite{Niemz2005}.

\begin{figure}
	\centering
	\includegraphics[width=\columnwidth]{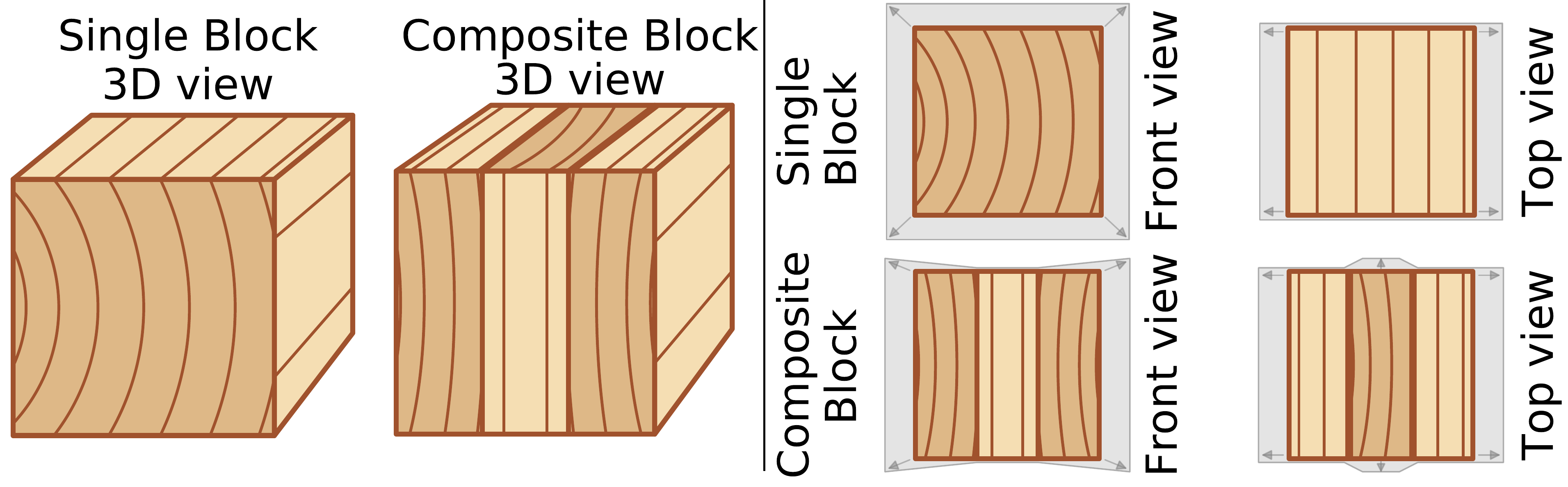}
	\caption{Humidity related swelling and shrinkage of wood is directional dependent (grain direction); Left: original state; Right: gray border indicates swelling effect}		
	\label{fig:Moosbrugger_Local_Global_ExpansionWrtHumidity}
\end{figure}

In order to homogenize these swelling and shrinking effects various wood-composite materials, like cross laminated timber, have been developed and are widely used today
\cite{Keylwerth1962,Keylwerth1968}.
Although these materials are more isotropic, strong local border effects still remain  \cite{Moosbrugger2017_T1,Moosbrugger2018_T3} (Figure \ref{fig:Moosbrugger_Local_Global_ExpansionWrtHumidity} ). 
Even though this is hardly a problem for a single composite block, these directional dependent local changes add up if a construction is assembled from  multiple composite blocks.

\begin{figure}[htbp]
	\centering
	\includegraphics[width=1.02\columnwidth]{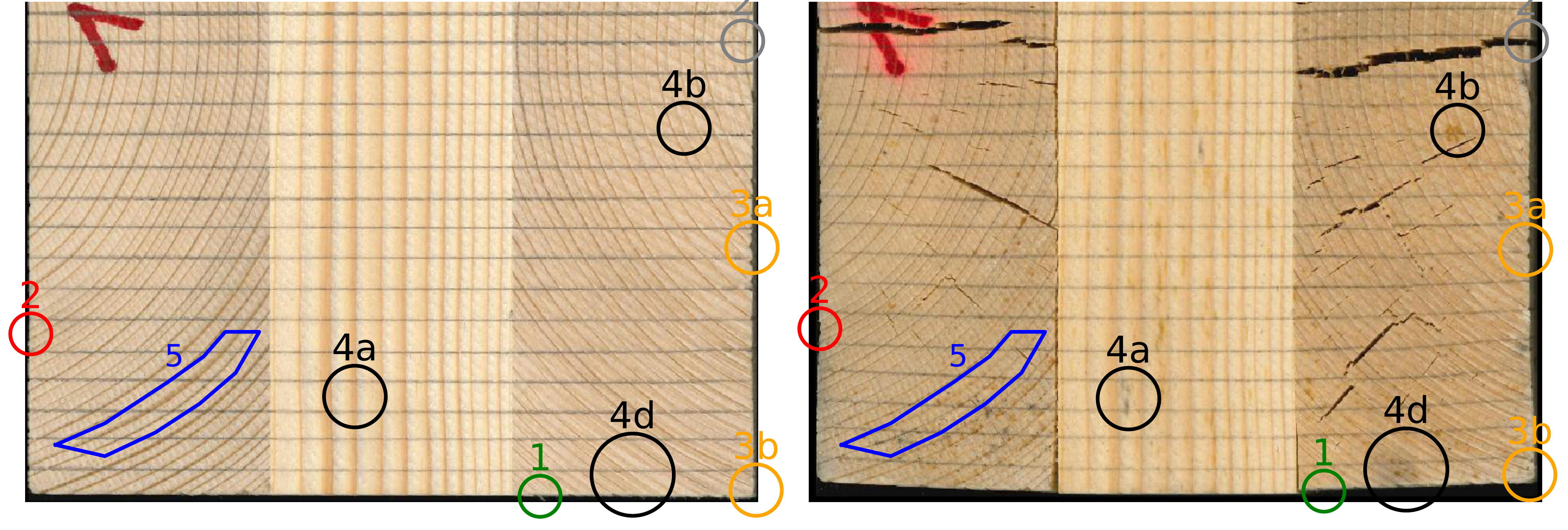}
	\caption{Image artifacts from the humidity conditioning process -- details only visible in digital version at high zoom; 1: artifact at the border, 2: damages to the border, 3: border not well defined due to rounding effects, 4: various stains from mold, 5: annular ring color change seems like shrinkage.}
	\label{fig:Problematic_Regions}
\end{figure}

To further improve these materials, as well as for the generation of key figures for construction and engineering, quantification of this deformation behavior is necessary.
Moosbrugger  et al. \cite{Moosbrugger2017_T1,Moosbrugger2018_T3} are currently conducting research to gather these coefficients for various materials, using the following procedure.
First, a specimen of the wood-composite under investigation is generated and both the starting humidity and key dimensions are measured manually.
Next, the specimen's humidity is altered in a controlled way which can take weeks.
Now all key positions are re-measured and deformations are calculated.
This process is repeated for all humidity values and specimens.
For documentation purposes high resolution images of the specimens faces are gathered at each step using a flatbed scanner  to reduce the projection error.
Figure \ref{fig:Problematic_Regions} shows cropped and aligned regions of such images in original and dry state.
Various changes like cracks, rounded and jagged borders, mold stains, color changes in the annular rings etc. can be observed.

\subsection{Contributions}
\label{sec:Intro_OurContribution}
In order to improve measurement accuracy, to increase the amount of calculated data-points and to reduce manual labor at the same time, we propose a novel approach based on computer vision that is fully automatic and can be used for a huge number of specimens, after an initial setup procedure. 

Our approach adds 3 automatic steps after image acquisition.
Those are,
 image preprocessing and alignment (sec. \ref{sec:Meth_ImagePreprocessing}), estimation of the optical flow and the brightness changes in the surface of the wood by means of optimization (sec. \ref{sec:Meth_OpticalFlow}), and finally calculation of the relative deformation field and 
the overall deformation coefficients (sec. \ref{sec:Meth_DeforamtionField}).

In a first 	step, a pre-processing pipeline automatically detects and calculates bounding boxes for the specimen in the images.
Then all the images showing the same face of the specimen at different humidity levels are aligned, cropped and saved as separate image files with same dimensions.

After that an image pair of the same face at different humidity levels is selected and the apparent motion of each pixel (optical flow) as well as additional brightness changes between the two images are estimated.
It is done using a variational optimization approach with Huber regularization \cite{Chambolle2010}.
This allows to specify how smooth or piecewise constant the final flow and illumination fields shall be estimated, which allows specification on how well cracks and surface color changes shall be modeled.

Finally, this flow field is used to calculate the relative deformation field in a dense manner for the whole 2D image as well as the desired engineering coefficients and their variation throughout the block.
The method is also robust as it reduces the effects from single artifacts (Figure \ref{fig:Problematic_Regions}), as the whole 2D image is taken into account in the optimization process.

\subsection {More simple but non robust alternatives}
We also investigated simpler ideas that turned out less accurate and robust.
However, with perfect conditions they yield reasonable results and have therefore been used to verify our main method (sec. \ref{sec:Exp_Comparison}).

Estimation of the coefficients using  automatically generated and matched image descriptors like SIFT, BRIEF or ORB \cite{scikit-image} does not work well due to the partly extreme brightness changes on the objects due to mold, minor damages, etc..
Especially in the case of dried wood blocks with cracks, standard feature detectors yield many false matches.
Furthermore, these descriptors are typically not sub-pixel accurate, whilst the optical flow estimation which serves as basis  for our main method has sub-pixel accuracy.

Deformation estimation that simply uses the blocks edge-boundary seems tempting as well. 
However, the edge-boundary is not well defined in the image, especially in the deformed cases where the flatbed scanner might also capture parts of the perpendicular face.
Furthermore the edges are only initially sharp and easily  get damaged due to handling and cracks as can be seen in Figure \ref{fig:Problematic_Regions}.

\subsection {Related Work}
\label{sec:Intro_RelatedWork}
Clocksin et al. \cite{Clocksin1997NewFlow,Clocksin02inspectionof} also use optical flow for inspection of surface strain in  materials like metals or ceramics.
They use a probabilistic inspired model to estimate a flow field that can model discontinuities and use it to estimate the overall small strain tensor using a least squares fit.
Their method is well suited for applications such as mechanical testing rigs with a fixed camera that continuously retains images.

Works by Rodriguez et al. \cite{Rodriguez2011} use a block-matching approach for a similar setup.

However, both approaches do not model surface color changes, nor do they cope with image registration, which both are absolutely necessary for our 
application.
In contrast, our approach allows to specify how piecewise smooth or constant our flow and illumination fields shall be estimated, which can be tuned for the needs of any application scenario.
Furthermore, we give a dense pixel-wise estimate of the surface strain and calculation procedures for the desired engineering coefficients including their variation throughout the block.

\section{Method}
\label{sec:Method}

In this section the complete method will be explained.
It starts with the preparation of the specimen and the gathering of the images followed by the aforementioned preprocessing step.
Next, the optical flow field will be estimated and finally the deformation characteristics are computed based on it. 

\subsection {Preparation of the specimen and acquisition of images}
\label{sec:Meth_ImageGathering}
The first part of the method consists in preparation of the specimens.
These wooden blocks consist of various plates that are laminated together  in  different orientations.
Then they are cut to given starting dimensions which ensures an approximate rectangular shape.
For the original manual measurement procedure of \cite{Moosbrugger2018_T3}, also various key points need to be marked.
Finally, the humidity is measured and images of the specimens are acquired using a flatbed scanner.
Since the specimens' surfaces are  planar in the beginning, their faces will align parallel to the scanner which keeps the projection error negligible.

Next, the specimens are put into a special environment for humidity control.
They are stored for multiple days up to weeks until a final homogeneous humidity level is reached throughout the block.
Finally, the current humidity is measured and  high-resolution scans are taken again. 

\begin{figure}[bh]
\centering
\includegraphics[width=0.9\columnwidth]{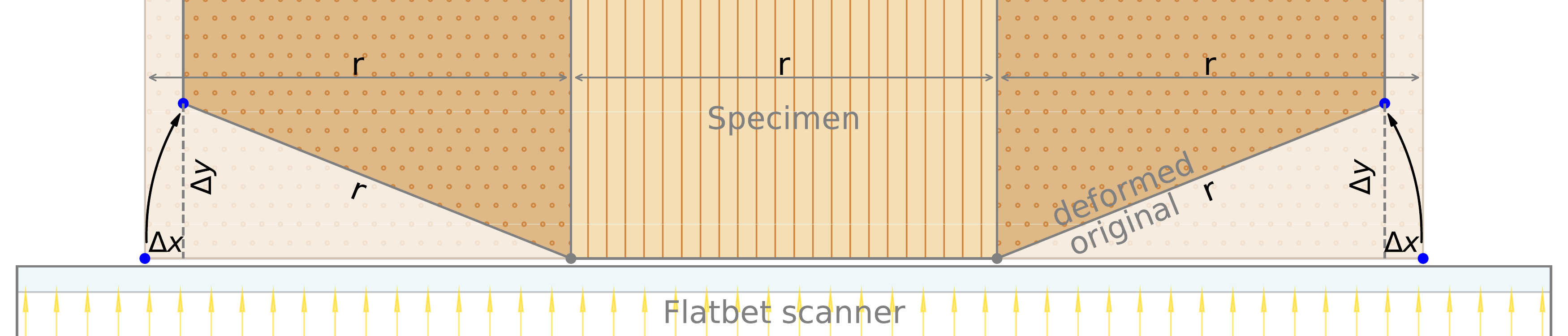}
\caption{Error estimation due to orthogonal changes}
\label{fig:ErrorEstimation}
\end{figure}

Since the specimens are now deformed their surfaces are not necessarily planar anymore.
Therefore, length changes on one side $\Delta y$ can lead to an apparent motion $\Delta x$ on the perpendicular side as well. 
Figure \ref{fig:ErrorEstimation} illustrates this, modeled with the  assumption that perpendicular changes can lead to shearing, which can be used to estimate the error as
\begin{equation}
\mathtt{err_{proj,max}} = r \left(1-\cos\left(
\arcsin \left (\frac{ \Delta y}{r} \right) \right) \right)
\end{equation}

For the specimen named PK1 of \cite{Moosbrugger2018_T3} the size of one segment is $r = 50$ mm, with a maximum reported length change of $\Delta y = \pm 4$ mm, which corresponds to $\Delta x = \pm 2$ mm that are visible for one face.
Therefore, the estimated maximum projection error is $\approx \pm 0.04$ mm which is around $\pm 1$ \% of the maximum perpendicular length change. 
This is an effect that may be observable in the final results, but is still in an acceptable range for this measurement purpose.

\subsection {Image preprocessing and alignment}
\label{sec:Meth_ImagePreprocessing}
\begin{figure*}[t]
	\centering
	\includegraphics[width=0.95\textwidth]{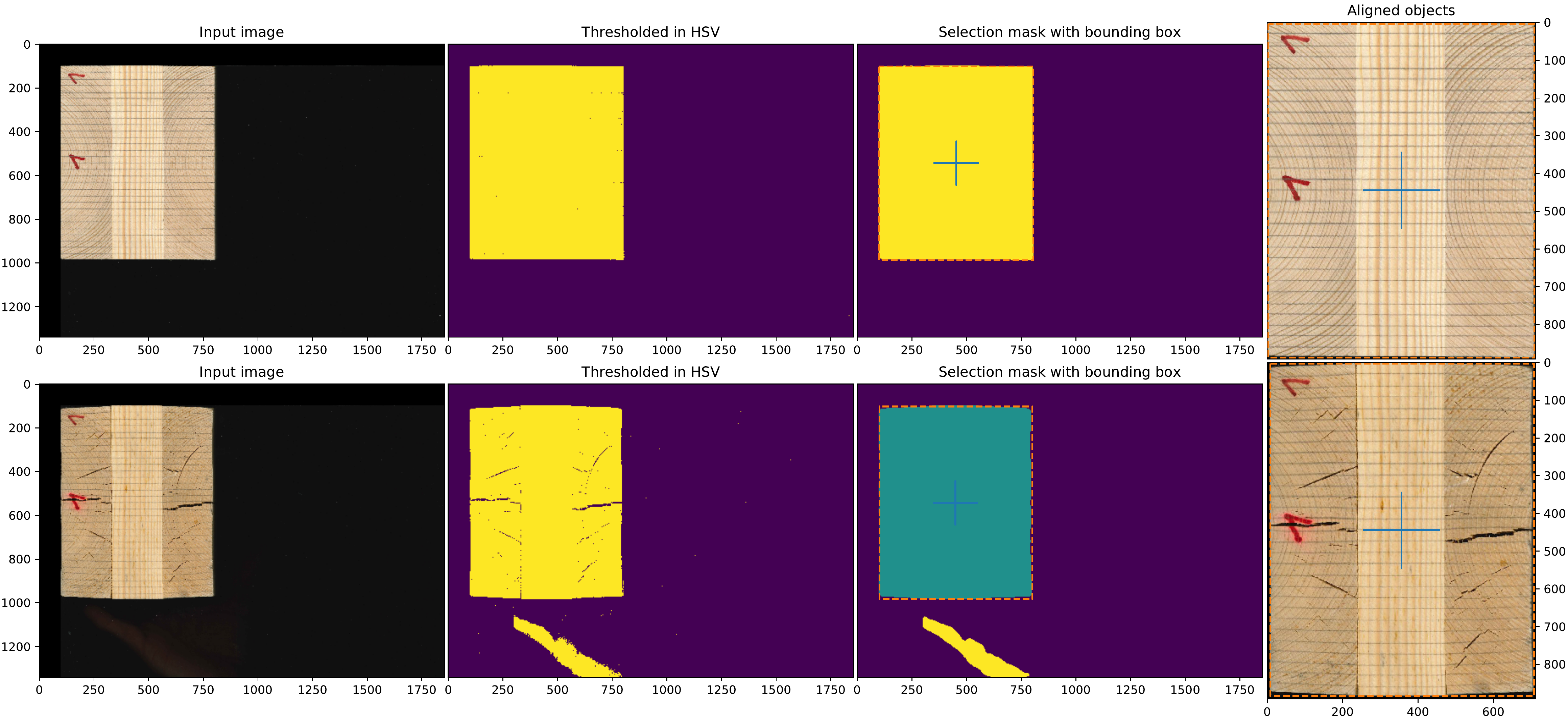}
	\caption{Illustration of the automatic image registration process. Top: $\mathtt{RH_0}\approx $ 12\% humidity, bottom $\mathtt{RH_1}\approx $ 25\% humidity. Left to right: Input image; automatic thresholding in HSV (For better illustration of the concept a non-ideal threshold has been chosen here); closed boundaries and object of interest selection; final aligned images; }
	\label{fig:Alignment}
\end{figure*}

Once the images of the specimen are available  at different humidity values, a multi step preprocessing procedure is conducted to prepare the images for optical flow calculation (Figure \ref{fig:Alignment}).
The main idea is to segment the object of interest - the face of the specimen - in all the images and to align the images showing the same face in different humidity states so that the geometrical center of the specimens face is well aligned over all the  images.

There are many possibilities on how the alignment may be achieved - the approach used in this work relies on generation of binary object masks and analysis of their properties.
For many parts of the implementation the python based image processing library scikit-image \cite{scikit-image} was used.
 An overview of the procedure will be given in the following.

First, the image resolution is reduced to a working resolution of 150 dpi and a black background border is added to ensure that none of the specimens faces is touching the border, which eases later processing.
Next, the color space is transformed from RGB to HSV (Hue Saturation Value) where the channels are less correlated and the object is easier to segment in the value.
Then the image is converted to black and white using a threshold that is automatically generated using Otsus method \cite{scikit-image}, which tries to maximize the variance for both black and white pixels.
Small errors are removed using a median filter.

In order to find the main object of interest the connectivity of white pixels is analyzed and the same label id is given to regions of pixels if they are connected by least one of their 8 surrounding  pixels (hor., vert., diag.).
The region connected to the  previously added border is the background that typically fills most of the image.
To fill up holes inside the objects, the labeling process is repeated for all the pixels that are not part of this connected background.
Now the amounts of pixels in each region are counted and the one with the highest count that is not the background is selected as the region of interest, using the white pixels as binary mask.

For this mask the dimensions are calculated and the center of gravity (CoG) is estimated using image moments.
The mask from the initial state where the specimen is still rectangular, is de-rotated to align well with the $x$ and $y$ axis so that it can serve as a reference.
The relative rotation of the other humidity states can later be estimated using the optical flow results.
This whole process is repeated for all the images of the same face of the specimen.

Finally, all images with the same face are aligned on their CoG and cropped according to the maximum $x$ and $y$ extent of the combination of all binary masks.
Each aligned image is stored separately as well as all the binary masks.

Since the specimen is not perfectly rectangular, especially in the deformed condition , there is a chance for higher projection errors at the border of the specimens face.
Furthermore, there are numerous artifacts that can happen at the border as illustrated in Figure \ref{fig:Problematic_Regions}.
Therefore, a data-term mask is also being generated that will be used in the optimization process to specify regions to ignore.
It is generated by shrinking the binary image mask using erosion with a diamond stencil in order to preserve sharp corners in the mask.


\subsection {TV-L$_1$ Optical Flow with compensation}
\label{sec:Meth_OpticalFlow}
Given the aligned images, the computation of the apparent motion of the pixels (optical flow) can be started.
The underlying idea is that the brightness between the two images remains similar during the motion, which is known as the brightness constancy assumption \eqref{eqn:OF_Brightness_consistency}.
Taylor approximation directly leads to the optical flow constraint, given here in scalar \eqref{eqn:OFC_traditional_scalar} and vectorial form  \eqref{eqn:OFC_traditional_vec}.
In this  notation $v=(v_x,v_y)^T$ describes the flow as vector field $v: \Omega \mapsto \mathbb{R}^2$,  $\nabla I$ is the derivative w.r.t. space and $I_t$ w.r.t. time.
\begin{eqnarray}
 \label{eqn:OF_Brightness_consistency}
I(x,y,t) &
\approx & I(x+\Delta x, y + \Delta y, t + \Delta t) \\
\nonumber
&
\approx 
& 
I(x,y,t)  + \frac{\partial I}{\partial x} \Delta x + \frac{\partial I}{\partial y} \Delta y + \frac{\partial I}{\partial t} \Delta t 
%
%
\\
%
 \label{eqn:OFC_traditional_scalar}
 0 &\approx &
 \frac{\partial I}{\partial x}  \underbrace {\frac{\Delta x}{\Delta t} }_{v_x}
 + 
 \frac{\partial I}{\partial y}  \underbrace {\frac{\Delta y}{\Delta t}  }_{v_y}
 + 
 \frac{\partial I}{\partial t}
 \\
 \label{eqn:OFC_traditional_vec}
 \rho(v) &=& (\nabla I)^T v + I_t
\end{eqnarray}
In practice, TV-L$_1$ flow estimation is repeated multiple times for each stage of a coarse to fine network, always using the flow estimate $v^0$ from the previous run to warp the target image $I_2$ incrementally closer towards the original $I_1$, which allows to estimate the flow incrementally finer, but requires re-computation of the gradients $I_t$ and $\nabla I$ in  each step.  

In our case the illumination can vary, which can be modeled by an additive scalar field $u: \Omega \mapsto \mathbb{R}$ \cite{Chambolle2010}.
Its overall influence can be specified with the scaling parameter $\beta$.
This results in the modified version 
\begin{eqnarray}
\rho_\beta(u,v) = I_t + (\nabla I)^T  (v -v^0) + \beta u
\end{eqnarray}
which is the final model describing our data-term.
To perform the optimization, a minimization approach that minimizes the total variation in the fields defined by the data-term is used, as described by \cite{Chambolle2010}.
It is based on minimizing the following variational model
\begin{eqnarray}
\label{eqn:OF_min_functional}
\min_{u,v} \| \nabla u\|_{\mathtt{1}} + \| \nabla v\|_{\mathtt{1}} +  \lambda \|  \rho_\beta(u,v) \|_{1} 
\end{eqnarray}

\begin{figure}[htbp]
	\includegraphics[width=0.95\columnwidth]{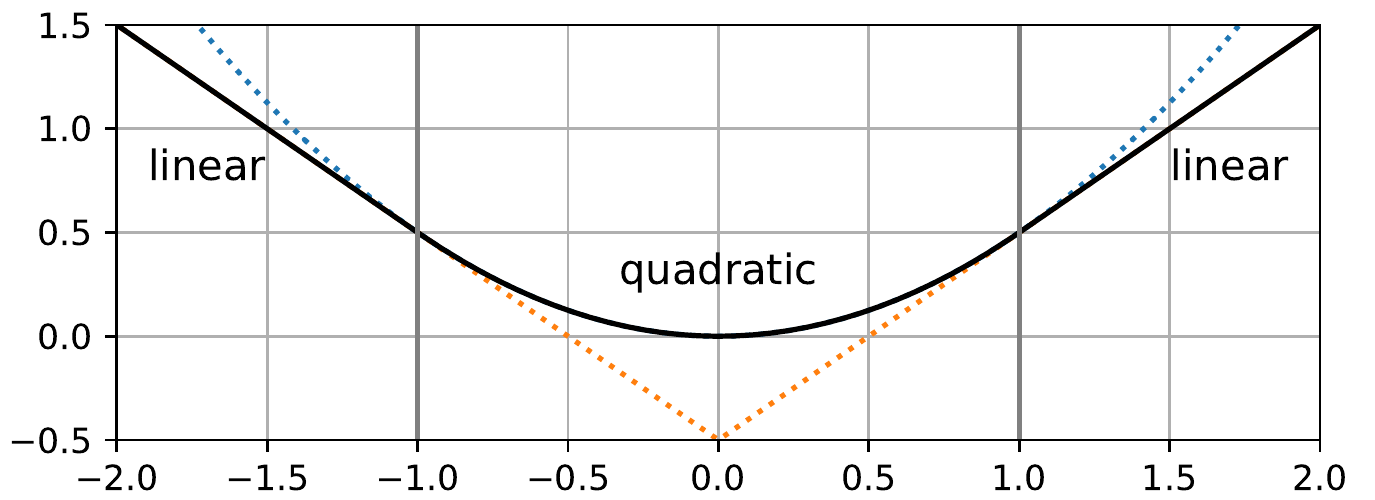}
	\caption{Huber norm with $\varepsilon_H$  = 1.0}
	\label{fig:Huber_norm}
\end{figure}

The terms $ \| \nabla u\|_{\mathtt{1}}$ and $\| \nabla v\|_{\mathtt{1}}$ in \cite{Chambolle2010} are the regularization terms that use a $L_1$ norm on gradients of the fields.
This regularization penalty results in fields that are piecewise constant.
For our case, we want a trade-off between mostly smooth fields -- which could be achieved using a $L_2$ norm -- since the flow and the illumination are expected to be mostly smooth, and also allow some discontinuity in the fields if the errors would otherwise be large. 
This is necessary to model local effects like cracks inside the blocks that correspond to discontinuities  in the flow field and to model mold and small stains that correspond to piecewise constant regions in the illumination as can be seen in Figure \ref{fig:Problematic_Regions}.
This can be achieved using a norm that uses the Huber function $\| \cdot \|_{\mathtt{H}}$ \cite[page 62 contd.]{Werlberger_convexapproaches} given by
\begin{eqnarray}
\label{eqn:Huber_norm}
\| p \|_{\mathtt{H}} &=& \sum_{i=1}^{N} h_{\varepsilon_H}(p_i)\\
h_{\varepsilon_H}(\alpha)  &= &
\begin{cases}
\frac{\alpha^2}{2 \varepsilon_H} & \mathtt{  if   } \; |\alpha| \leq \varepsilon_H\\
|\alpha| - \frac{\varepsilon_H}{2}
\end{cases}
\end{eqnarray}
 and illustrated in Figure \ref{fig:Huber_norm}. 
It basically models a smooth transition from a $L_2$ norm to a $L_1$ norm that can be chosen with the parameter $\varepsilon_H$

Since we also have regions that show invalid data, like the border and the outside regions, we want to have the possibility to mask out pixels that are not of interest and better ignored in the optimization process.
This can be achieved using the aforementioned data-term-mask, named $m$ here, which is multiplied element-wise (Hadamard product $\odot$)  with the data-term.

This leads to our final variational model 
\begin{eqnarray}
\label{eqn:OF_min_functional_final}
\min_{u,v} \| \nabla u\|_{\mathtt{H,ilu}} + \| \nabla v\|_{\mathtt{H,Flow}} +  \lambda \| m \odot   \rho_\beta(u,v) \|_{1}
\end{eqnarray}
which is optimized using the  primal dual algorithm \cite{Chambolle2010}.
At this point only an intuition of this algorithm will be given, for implementation details please refer to \cite{Chambolle2010}.
The idea is to convert the primal minimization problem into a saddle point problem where optimization takes place by minimizing the primal and maximizing the dual variables in each iteration which has speed-up benefits for this kind of non-smooth optimization problem.

The rotation between the two images can be estimated with the difference in the angle $\Delta \theta$  formed by the original  $x,y$ coordinates angles and angles from the deformed coordinates $\tilde{x}, \tilde{y}$.
Averaging over $v$ and $\Delta \theta$ gives an estimate for the average displacements $v_{\text{avg}}$ and average rotation $\Delta \theta_{\text{avg}}$.
Optionally, $ \Delta \theta_{\text{avg}}, v_{\text{avg}}$ can be used to further optimize the registration of image $I_2$ w.r.t. $I_1$ in combination with a repetition of the whole flow estimation process.
\begin{eqnarray}
\tilde{x} 
\!\!&=&\!\!
 x + v_x (x,y)
\hspace{1cm}
\tilde{y} = y + v_y (x,y)  
\\
\Delta\theta
\!\!&=&\!\!
\tilde{\theta} - \theta =
\arctan \left( \frac{ \tilde{y} }{ \tilde{x}} \right) - \arctan \left(   \frac{y}{x} \right)    
\\
\!\!&=&\!\!
 \arctan \left( \frac{ \sfrac{\tilde{y}}{\tilde{x} } - \sfrac{x}{y} } { 1 + \sfrac{\tilde{y}}{\tilde{x} } \;  \sfrac{x}{y} } \right)
\end{eqnarray}

\subsection {Calculation of the deformation field}
\label{sec:Meth_DeforamtionField}
In this interdisciplinary section the exemplary calculation of one relative deformation coefficient for engineering will be shown.
It is done by relating strain tensors known from classical mechanics of materials \cite{mang2013festigkeitslehre} with our flow field.
These tensor fields are then used to calculate the engineering coefficient in a full and a more simplified version.

The engineering coefficient on which we demonstrate the procedure is $k_{cor,a}$ from \cite{Moosbrugger2018_T3}\eqref{eqn:Deform_orig_xkor} that we shortly call $k$ here.
It describes the relative length change w.r.t. the humidity change  and is calculated independently for each side of the block.
This is basically a global estimate for the normal strains as seen at the surface w.r.t. to the humidity change.
In \cite{Moosbrugger2018_T3} it is calculated using manual caliper measurements of the lengths in original $l_{0}$ and deformed state $l_{1}$ together with the according relative humidities $\mathtt{RH}_i$.
\begin{equation}
\label{eqn:Deform_orig_xkor}
k = \frac{(l_0 - l_1) }{l_0 (\mathtt{RH}_0-\mathtt{RH}_1)}
\end{equation}

In mechanics of materials \cite{mang2013festigkeitslehre}, strains fields are described by the Green strain tensor $E$ and the small strain tensor $e$.
The strain tensor $e$ is  a linearized version of E and is used more widely in engineering due to its simplicity,  although it is less accurate.
\begin{eqnarray}
E =
\begin{bmatrix}
E_{11} 	& E_{12} \\
E_{21}  & E_{22} 
\end{bmatrix} 
\hspace{0.5cm}
\approx
\hspace{0.5cm}
e =
\begin{bmatrix}
\epsilon_{11} 	& \gamma_{12}/2 \\ 
\gamma_{21}/2  & \epsilon_{22}  \\
\end{bmatrix}
\end{eqnarray}
The principal diagonal elements of these tensors are related with the normal strains termed $\epsilon$  whereas the other elements $\gamma_{ij}$ and $E_{ij}$ correspond to shearing  effects.
These strains can be calculated by derivation of the absolute displacement vector which in our case corresponds to the flow field.
In our notation $\epsilon_{ii}$ is based on the small strain, whereas $\epsilon_i$ is based on the Green strain.
The coordinates $x$ and $y$ always correspond to the non-deformed state of the object.
\begin{eqnarray}
\begin{matrix}
\epsilon_{11} = \frac{\partial v_x}{\partial x},
&
\epsilon_{22} = \frac{\partial v_y}{\partial y},
&
\gamma_{12} = \gamma_{21} =  \frac{\partial v_x}{\partial y} + \frac{\partial v_y}{\partial x} 
\end{matrix}
\\
\begin{matrix}
\epsilon_1 = \sqrt{ 1+ 2 E_{11} }-1
&, &
\epsilon_2 = \sqrt{ 1+ 2 E_{22} }-1
\end{matrix}
\\
E_{11}(x,y) = 
\sfrac{\partial v_x}{\partial x} 
+
\frac{1}{2} 
\left[
\left(  \sfrac{\partial v_x}{\partial x}  \right) ^ 2
+
\left(  \sfrac{\partial v_y}{\partial x}  \right) ^ 2
\right]
\\
E_{22}(x,y) = 
\sfrac{\partial v_y}{\partial y} 
+
\frac{1}{2} 
\left[
\left(  \sfrac{\partial v_x}{\partial y}  \right) ^ 2
+
\left(  \sfrac{\partial v_y}{\partial y}  \right) ^ 2
\right]
\end{eqnarray}

Normalization of the normal strains with the humidity change that caused the deformation results in a dense estimate for  $k$ in form of a point wise vector field with the $x$ and $y$ component given by
\begin{eqnarray}
k_{x}(x,y) =  \frac{ \epsilon_{1}(x,y)}{\Delta \mathtt{RH}} \approx  \frac{ \epsilon_{11}(x,y)}{\Delta \mathtt{RH}} =  \frac{\partial v_x(x,y)}{ \partial x  \;\;  \Delta \mathtt{RH}} 
\\
k_{y}(x,y) =  \frac{ \epsilon_{2}(x,y)}{\Delta \mathtt{RH}} \approx  \frac{ \epsilon_{22}(x,y)}{\Delta \mathtt{RH}} =  \frac{\partial v_y(x,y)}{ \partial y  \;\;  \Delta \mathtt{RH}} 
\end{eqnarray}

\begin{figure}[htbp]
	\includegraphics[width=0.95\columnwidth]{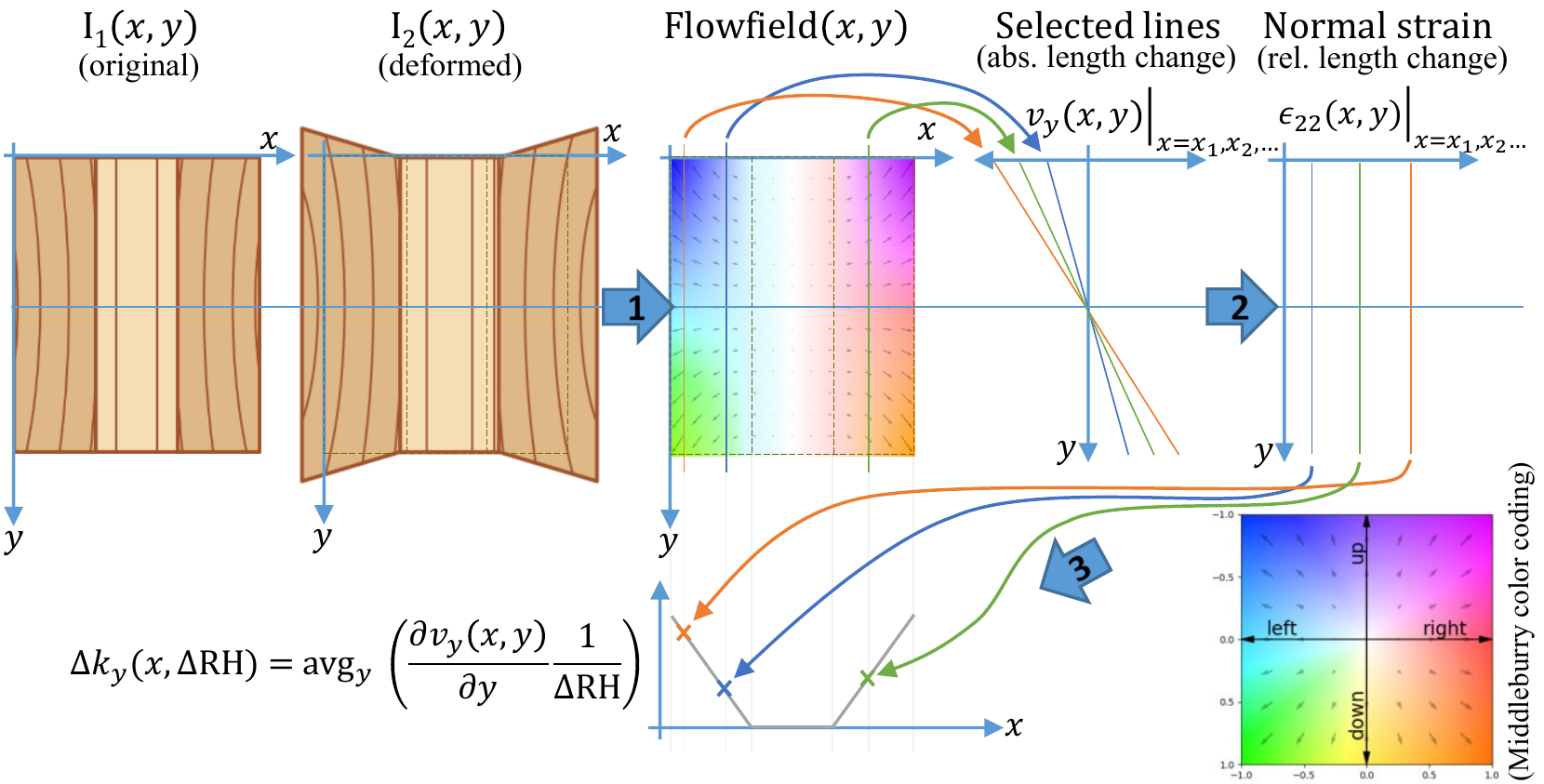}		
	\caption{Conceptual illustration of the whole method. 1: Flow estimation for an image pair; $y$ deformation $v_y$ is shown for selected $x$ positions; 2: Derivation w.r.t. $y$ leads to $\epsilon_{22}$; 3: Averaging over $y$ and normalization yields $k$}
	\label{fig:MethodConcept}
\end{figure}

If the block would be perfectly homogeneous per direction, every normal-strain $\epsilon_{22}$ along the $y$-axis of a given $x$-position would be equal to the overall normal-strain at the same $x$-position.
This can be seen in Figure \ref{fig:MethodConcept} that illustrates the overall concept.

However, in a real specimen the normal-strain changes throughout the block, due to inhomogeneities such as annular rings, cracks etc.
The resulting global normal-strain can therefore be estimated by averaging.
But since we are interested in how the $y$ dimension changes over $x$, the averaging is only done along the $y$ dimension \eqref{eq:kcor_y_BothStrains}.
The same can be done vice versa for the $x$ dimension changes \eqref{eq:kcor_x_BothStrains}.
The averaging range is set according to the data-term mask where the flow was optimized for.
\begin{eqnarray}
\label{eq:kcor_y_BothStrains}
k_{y}(x) &=&  \text{avg}_y \left( \frac{\epsilon_{2}}{\Delta \mathtt{RH}}  \right) \approx \text{avg}_y \left( \frac{\epsilon_{22}}{\Delta \mathtt{RH}}  \right)
\\
\label{eq:kcor_x_BothStrains}
k_{x}(y) &=&  \text{avg}_x \left( \frac{\epsilon_{1}}{\Delta \mathtt{RH}}  \right) \approx \text{avg}_x \left( \frac{\epsilon_{11}}{\Delta \mathtt{RH}}  \right)
\end{eqnarray}
If the block has cracks, extreme strain values appear at the discontinuities, which  can be used to find the cracks. 
Although, this increases the variance of $k$, the value for the small strain version $k_e$ reflects the overall value as seen at the averaging boundary ($a$ and $b$) ignoring inner distribution, due to a telescopic sum \eqref{eqn:ky_simplified} as averaging and derivation directions are aligned.
This is not the case for $k_E$ from the Green tensor.
Absolute strain values roughly an order of magnitude higher then the biggest small strain estimate for $k_e$ are considered cracks and therefore best omitted for the average and variance of $k_E$.
\begin{eqnarray}
k_{y,e}(x) 
&=& \frac{1 }{y_b - y_a} \sum_{y=y_a}^{y_b} \frac{ \epsilon_{22}(x,y)}{\Delta \mathtt{RH}}
\nonumber
\\
\label{eqn:ky_simplified}
= \dots &=&    \frac{v_{y}(x,y_b) - v_{y}(x,y_a)}{(y_b-y_a)\Delta \mathtt{RH}}
\\
k_{x,e}(y) &=&   \frac{v_{x}(x_b,y) - v_{x}(x_a,y)}{(x_b-x_a)\Delta \mathtt{RH}} 
\end{eqnarray}

To sum up, the correction factor $k$ can be calculated  for any desired region or direction of the block directly from flow field.
Furthermore, the block's strain tensors $e$ and $E$ are computed as dense fields, which can be useful for many applications.

\section {Conducted experiments}
\label{sec:Experiments}
Now that the method has been presented, various experiments will be made for setup, validation and demonstration purposes.

\subsection{Setting up the regularization and hyper-parameters for the flow}
\label{sec:Exp_Hyperparams}
The optimization step of our method contains hyper-parameters that need to be set up.
For some of these parameters rules of thumb can be given that work for wide ranges of data, but some others need to be adjusted more carefully to the current dataset.

For example, the results will generally be better the more warps, primal-dual-iterations, and pyramid layers are used, but values of around 5-10 warps, 30-100 primal-dual iterations, a pyramid scale of around 0.8-0.9 that corresponds to around 10-20 steps depending on image size will usually work fine.

Setting up the parameters that are associated with equation \eqref{eqn:OF_min_functional_final} is heavily dependent on the target data.
These parameters basically specify how piecewise-smooth we want our flow ($\varepsilon_{\mathtt{H,Flow}}$) and illumination ($\varepsilon_{\mathtt{H,ilu}}$) fields to be estimated, how big we want the impact of the illumination compensation to be ($\beta$) and how much influence the overall data-term will have in comparison to the regularization that implies the penalties on the variation in the fields ($\lambda$).

\begin{figure}[htbp]
	\includegraphics[width=0.90\columnwidth]{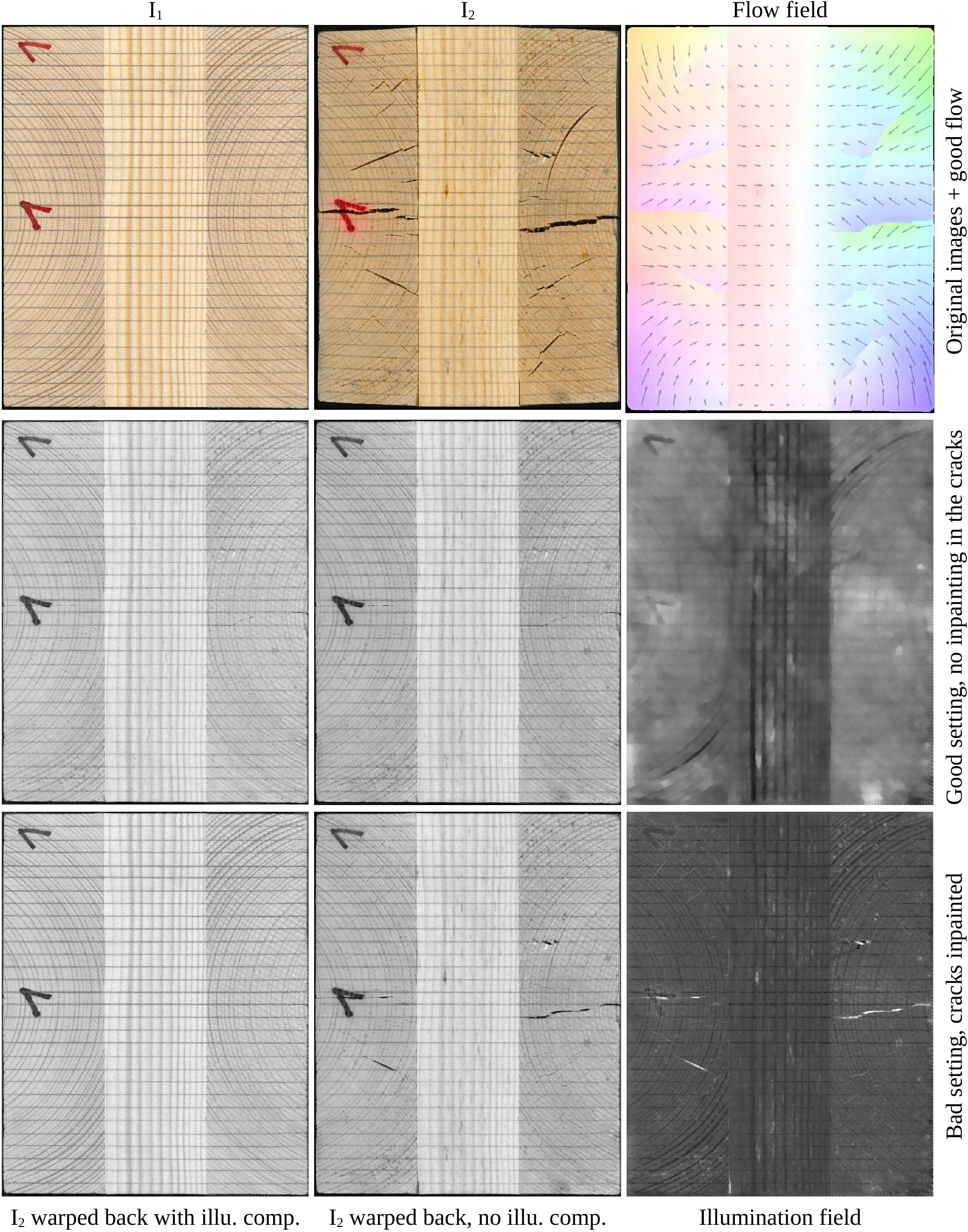}			
	\caption{Parameter setup: Good: flow alone closes the cracks; Bad: inpainting from illumination compensation }		
	\label{fig:Ex_good_setup}
\end{figure}

Before the setup procedure will be explained, an example of a good setup will be shown in Figure \ref{fig:Ex_good_setup} mid section.
This Figure was generated using 
$\varepsilon_{\mathtt{H,Flow}}$ = 0.2
$\varepsilon_{\mathtt{H,ilu}}$ =  0.05
$\beta$ = 0.04 
and
$\lambda$ = 10
.
Here the two lower leftmost images show the final warped version of $I_{2,w}$ with and without illumination compensation.
As can be seen the algorithm managed to close the cracks just with flow information and the intensity compensation just dampened the effects of stains to some degree.
These parameters were acquired using the following iterative procedure.
First, setting $\beta$ = 0 to remove illumination compensation, whilst working regions for $\varepsilon_{\mathtt{H,Flow}}$ and $\lambda$ are being tested out.
This can be verified by comparing the final warped image $I_2,w$ to the original version of $I_1$.
Once this is achieved, the illumination compensation term can be activated by setting $\beta$ to some value $\leq$ 1 and specifying how piecewise constant/smooth the illumination compensation shall be using $\varepsilon_{\mathtt{H,ilu}}$.
However, since these parameters setup  competition weights in the optimization process, changing a single one leads to overall changes as well.
Therefore, the final setup needs to be done jointly and iteratively.
This setup is best done on a selection of image pairs from the dataset that require vastly different flow and illumination fields, to ensure that the settings will work well for all images.
For example, the image pairs should contain combinations with smooth and discontinuous (small, large stains)  brightness changes as well as no changes.
The same should be true for the flow, which should contain image pairs with no flow, smooth, discontinuous flow (cracks) etc.

Care must be taken to avoid that the intensity compensation models too fine details. 
This can lead to inpainting effects, as can be seen in Figure \ref{fig:Ex_good_setup} bottom, where the flow was not able to close the cracks, and the compensation simply re-painted it.

\subsection{Comparison with other methods}
\label{sec:Exp_Comparison}
\begin{figure}[htbp]
\includegraphics[width=\columnwidth]{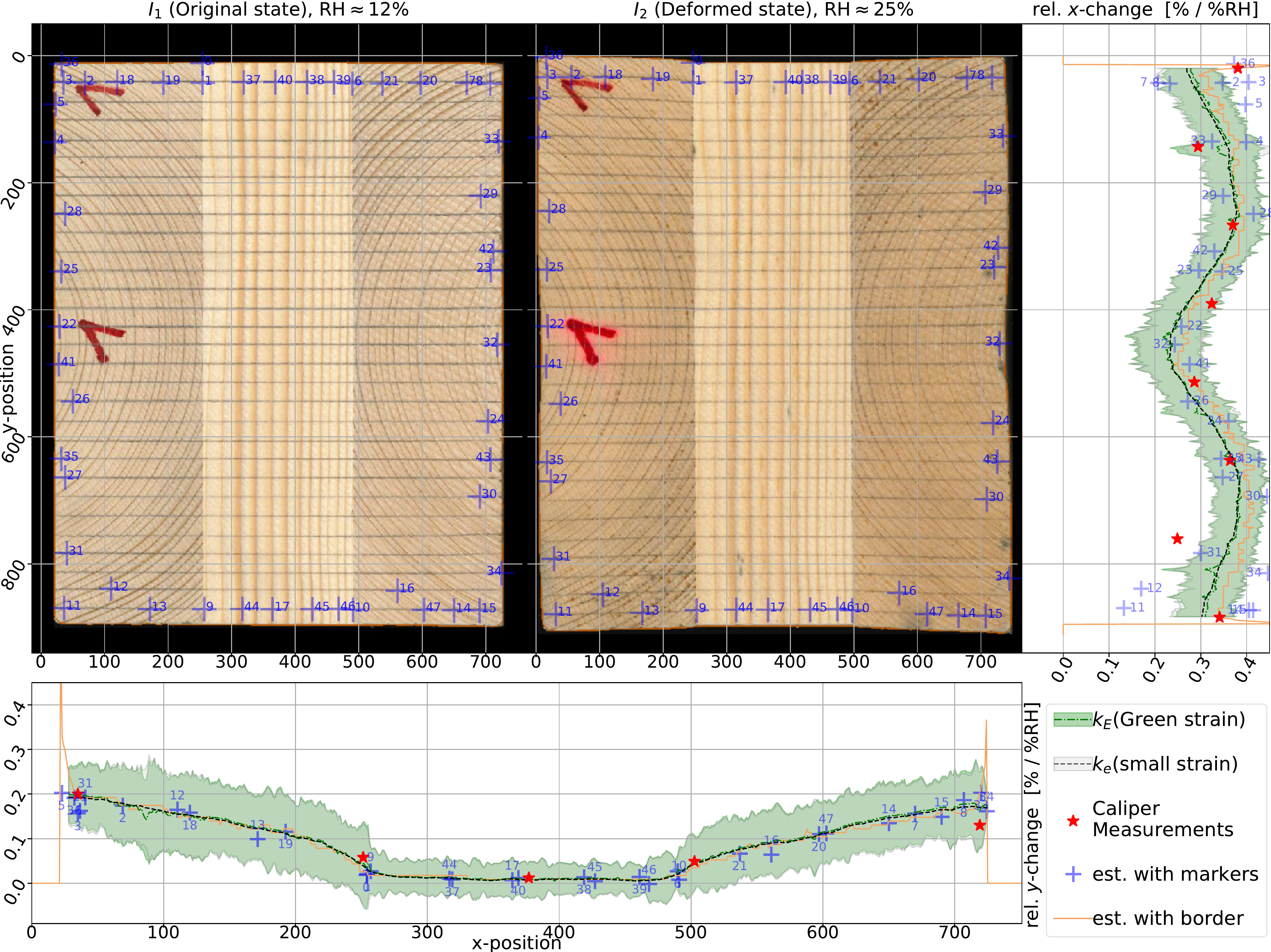}	
\includegraphics[width=\columnwidth]{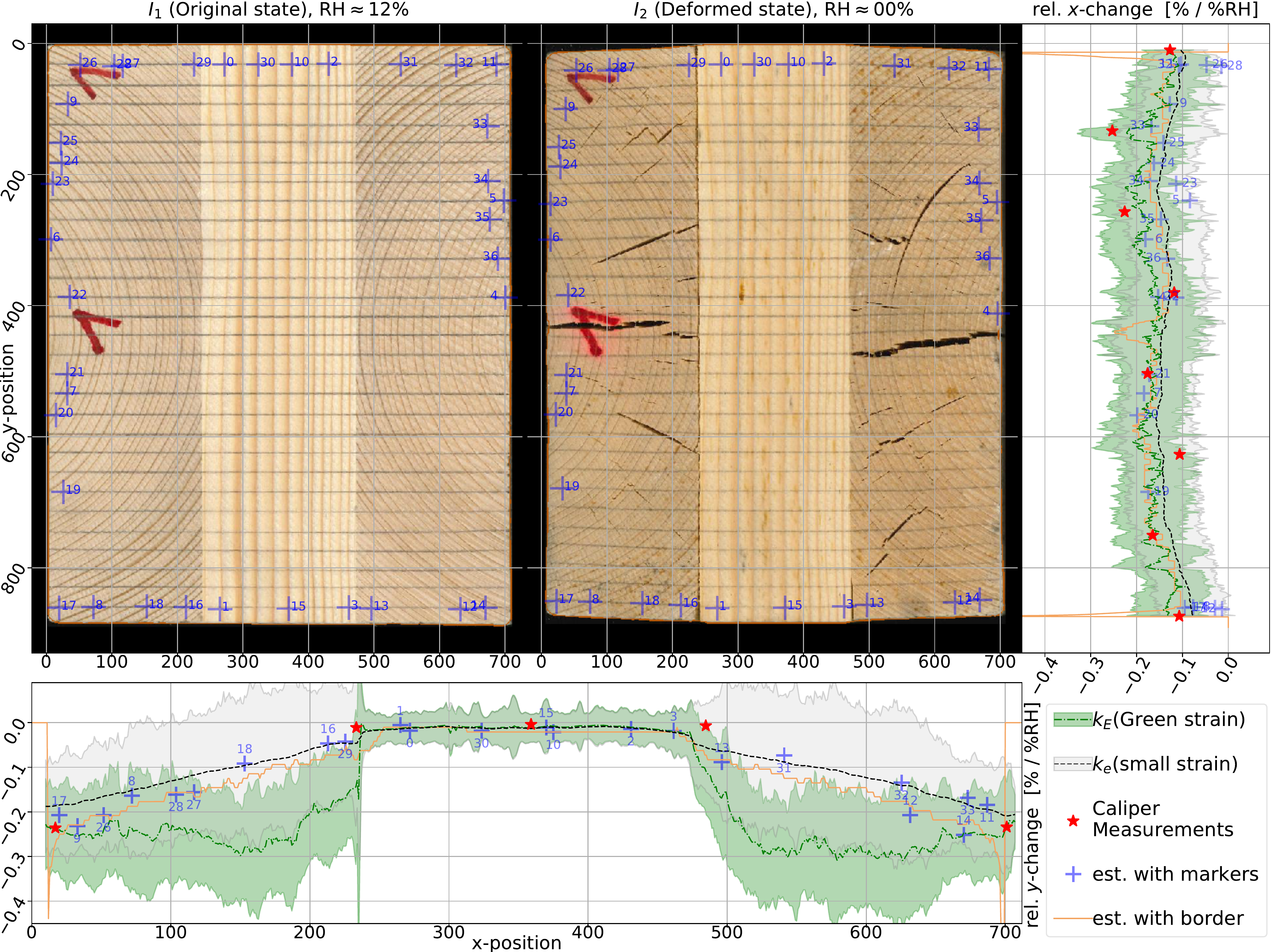}
	\caption{
Comparison - Rel. length change coefficient $k$: Estimates from small strain $k_e$ and Green strain $k_E$ allow predictions of variation in the block;
$k_E$ accounts for increased shrinkage (no material in cracks); Without cracks: $k_e \approx k_E$;  good matching with manual measurements
}	
	\label{fig:Ex_ComparisonWOtherMethods}
\end{figure}

In order to verify if the results of the main method are plausible, comparison against different measurement options have been carried out (Figure \ref{fig:Ex_ComparisonWOtherMethods}).
The first method uses manually placed markers on both images, that have been fine tuned using correlation.
All points have been visually checked after the correlation to avoid wrong correspondences due to stains mold etc.
Furthermore, only points that are sufficiently close to the border under investigation (for $y$ top-bottom) are taken into account to avoid local effects, as e.g. visible close to the annular rings.
$k$ is calculated according to equation \eqref{eqn:Deform_orig_xkor} for each point.
Since the front face of specimen PK1  shows a more or less well defined boundary in both images, also an estimation using the boundaries of the binary mask can be conducted in this particular case.
Here the lengths for  equation \eqref{eqn:Deform_orig_xkor}  are measured from the binary masks along the $x$ and $y$ direction respectively.

As can be seen the method with the markers, even though fine tuned with correlation, has a high variance on the output.
Also the method using the image mask's border has a high noise, even though the border is particularly well defined in this case.
For other specimen this method rendered completely useless. 
Our main method on the other hand yields a dense result for $k$ that aligns well with the noisy measurements from the other methods and it also allows to calculate the variance of the coefficient throughout the block along with the averaging process.
In the case of the dried specimen, the estimate for $k_E$ based on the Green tensor seems to account that the block actually became smaller than apparent on the surface due to the voids in form of the cracks.
On the other hand the small strain estimate for $k_e$ better reflects the values near the surface, due to the telescopic sum in the averaging process \eqref{eqn:ky_simplified}.

\section{Conclusion and future work}
As shown in experiments, our approach is able to extract humidity dependent strain properties directly from the images of the specimen at different humidity levels.
This works very robust despite the specimens' appearance-changes during the humidification process.
Our method describes exemplary estimation of one particular deformation coefficient and its variance.
However, since we estimate the whole strain field in a dense manner, various other coefficients might be estimated using this method as well.
Since our method allows accurate and separate setup of how piecewise constant or smooth the flow and illumination fields shall be estimated, it can easily be adapted to a variety of similar applications.

Future work includes investigations on expansion of the method to other useful engineering coefficients.
Furthermore, more in depth analysis on the absolute achievable accuracy using well defined test objects might be valuable.

\section*{Acknowledgements}
This work was supported by the research initiative Intelligent
Vision Austria with funding from the AIT and the Austrian Federal
Ministry of Science, Research and Economy HRSM programme
(BGBl. II Nr. 292/2012)


%
%




{\small
	\bibliographystyle{ieee}
	\bibliography{cvww_template}
}

\end{document}